\documentclass[runningheads]{llncs}

\usepackage{eccv}

\usepackage{eccvabbrv}

\usepackage{graphicx}
\usepackage{svg}
\usepackage{booktabs}
\usepackage{multirow}
\usepackage{amssymb}
\usepackage{xcolor}
\usepackage{bbm}
\usepackage{pifont}

\usepackage[accsupp]{axessibility}  

\usepackage{hyperref}

\usepackage{orcidlink}

\begin{document}

\title{CasaMaestro: Multi-View Panoramas for House-Scale 3D Reconstruction} 


\author{Yuzhou Ji\textsuperscript{*}\orcidlink{0009-0009-3572-060X} \and
Xiaotian Yang\textsuperscript{*}\orcidlink{0009-0009-4797-1254} \and
Zhipeng Zhang\textsuperscript{\ensuremath{\dagger}}\orcidlink{0000-0003-0479-332X}
}

\authorrunning{Y.~Ji et al.}

\institute{AutoLab, School of Artificial Intelligence, Shanghai Jiao Tong University, China\\
\url{https://george-attano.github.io/CasaMaestro}\\
\email{jiyuzhou@sjtu.edu.cn}, \email{zhipeng.zhang.cv@outlook.com}}

\maketitle

\begingroup
\renewcommand{\thefootnote}{\fnsymbol{footnote}}
\footnotetext[1]{Equal contribution. \ensuremath{\dagger} Corresponding author.}
\endgroup

\begin{figure}[h]
  \centering
  \includegraphics[height=5.5cm]{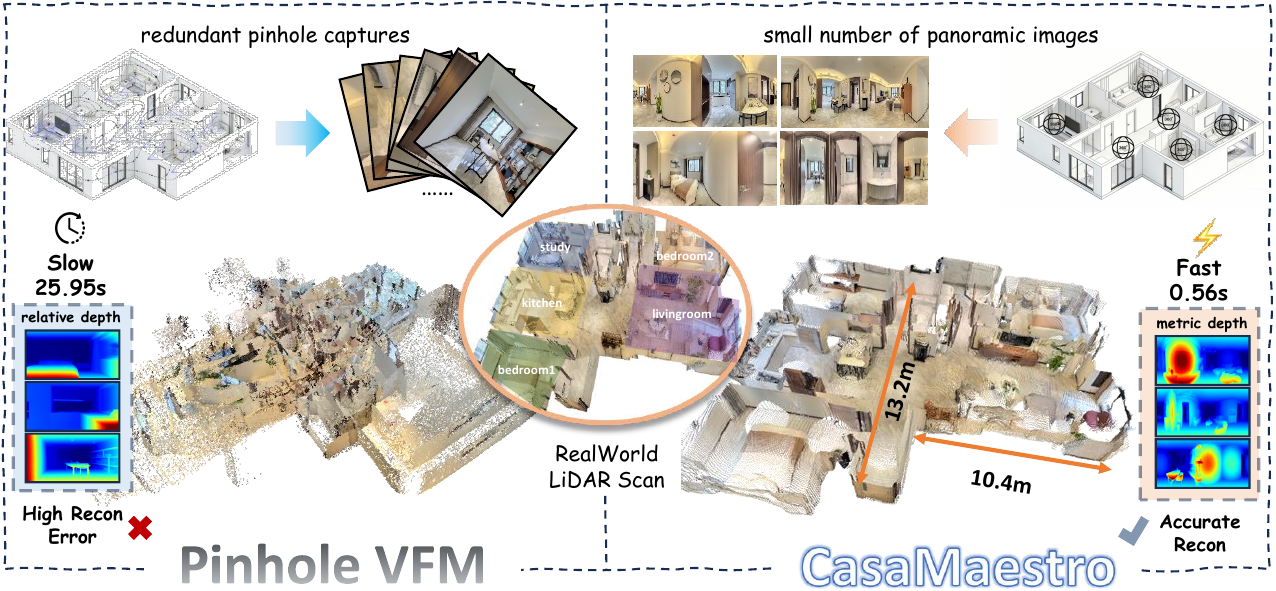}
  \caption{\textbf{Demonstration of CasaMaestro.} The existing pinhole vision foundation models (VFM) require dense video streams for scene-level reconstruction, which is consuming in both human effort and computing resources, while also facing severe errors in overly long and short input sequences. We propose CasaMaestro, which takes only sparse indoor panoramas and directly completes house-scale (with multiple rooms) metric reconstruction, providing efficiency along with accuracy.
  }
  \label{fig:teaser}
\end{figure}

\begin{abstract}
The rise of home-deployed embodied AI systems is driving a growing need for fast, metric 3D reconstruction of residential spaces to support navigation, interaction, and long-horizon task execution.
However, the commonly used pinhole-camera 3D reconstruction pipelines struggle to model large indoor residences efficiently due to their limited field of view, to which achieving full coverage across multiple rooms often requires thousands of images and incurs drift from long chains of incremental alignment.
In this work, we present CasaMaestro (\textit{Spanish} words meaning ``house'' and ``master''), a feedforward model that can take only twenty to fifty sparse multi-view indoor panoramas as input and directly predicts metric depth along with camera poses, allowing fast point-cloud reconstruction of the entire house with full coverage.
CasaMaestro is the first model that supports house-scale reconstruction with multi-view panoramas.
Experiments show that CasaMaestro can robustly provide high quality results in both real-world and synthetic scenes, which can serve as a strong foundation for acquiring house-scale 3D indoor assets to be applied in close-loop simulation.
\end{abstract}

\section{Introduction}
\label{sec:intro}

\begin{figure}[!t]
  \centering
  \includegraphics[height=3.20cm]{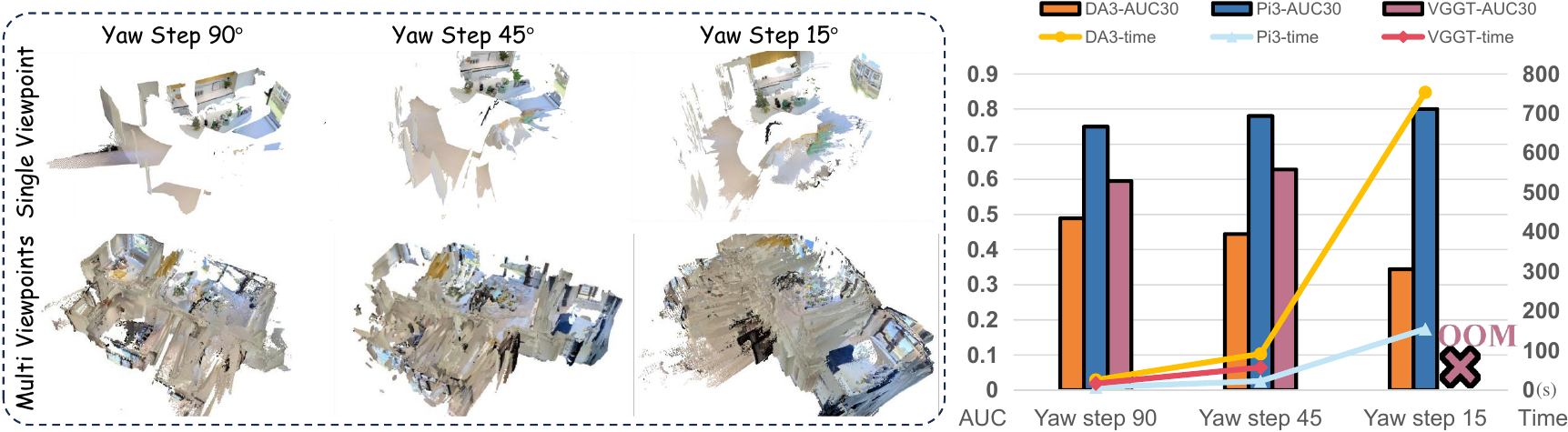}
  \caption{\textbf{Illustration of existing problems.} Left visualization shows pinhole models either face limited FoV in sparse capture or accumulative error in dense sequence. Right figure shows pose accuracy and processing time under different input density.}
  \label{fig:intro}
\end{figure}


With embodied AI shifting toward in-the-wild operation in everyday homes, the construction of realistic and metrically consistent closed-loop simulation environments becomes increasingly important.
In order to bridge the sim-to-real gap, reconstructing virtual 3D residential scenes from real-world capture has served as a crucial role for both training and testing, which, however, still faces many difficulties in practice.

Although LiDAR scanning remains a reliable solution for acquiring high-quality 3D assets of homes, it is costly and hardware-dependent.
In contrast, recent 3D vision foundation models \cite{depthanything3, wang2025pi3, wang2025vggt, keetha2025mapanything} suggest a compelling alternative by reconstructing 3D from unconstrained image sequences, enabling 3D lifting for existing and even synthetic imagery without specialized sensors.
However, today’s large-scale multi-view methods predominantly assume pinhole cameras, whose narrow field of view makes residential capture inefficient and fragile, where dense imaging is required for multi-room coverage and long trajectories exacerbate drift and error accumulation.
As shown in \cref{fig:intro}, pinhole methods suffer from severe displacement with sparse capture viewpoints that can not be solved (and may even worsen) by increasing single viewpoint scanning density due to accumulative error, while the time and GPU memory consumption are already extremely high.
This fact substantially prevents the practical deployment of pure-vision reconstruction for real-home data acquisition.

To improve reconstruction efficiency, several researches have explored using panoramic cameras, yet they are mostly restricted to single-view depth prediction \cite{lin2025depth, li20252} or to small-scale, short-range multi-view settings.
For example, PanoSplatt3R \cite{ren2025panosplatt3r} provides feed-forward 3DGS reconstruction from only panoramas, but can only handle 2 input views.
PanoPose \cite{tu2024panopose} uses separate pose network dedicated for relative pose estimation of panorama pairs to progressively predict long sequences, yet requires large view overlap and consequently a long image sequence for house-level scan.
Although SPR \cite{zheng2025scene} supports relatively larger view displacement comparing with PanoPose, it still demands video streams and is solely trained on 5 views setting, leaving the challenge of house-scale multi-view panoramic reconstruction unresolved.


In this work, we present CasaMaestro, the first feedforward model that achieves extrinsic-free multi-view panoramic 3D reconstruction with sparse house-scale capture. Embracing a minimalist design philosophy \cite{depthanything3}, CasaMaestro is simply built upon a DINOv2 \cite{oquab2023dinov2} backbone that processes multiple views, paired with dedicated heads for depth and pose prediction. To explicitly tackle the severe view displacements inherent in sparse residential captures, we introduce a lightweight panoramic camera pose decoder. By incorporating a second stage attention mechanism across views, this decoder achieves superior pose estimation accuracy. Moreover, recognizing that existing datasets offer a restricted variety of scenes and viewpoints, we propose a novel panoramic data augmentation strategy via ERP (Equirectangular Projection) remapping, which generates abundant pairs of poses and views, significantly boosting model robustness. Ultimately, CasaMaestro takes only raw panoramas as input to directly predict metric depth and camera poses. This elegant pipeline is remarkably simple yet surprisingly effective, which enables the immediate point cloud reconstruction of entire homes, delivering full coverage and exceptional geometric consistency.

Experiments show that our model is superior than existing methods with specifically \textbf{84\%} and \textbf{119\%} improvements of the AUC30 metric in real-world and synthetic scenes with balanced rotation and translation error, and also an average \textbf{21.98\%} decrease of AbsRel on unseen datasets comparing with the best previous results.

In conclusion, we provide the following contribution:

\begin{itemize}
    \item We present CasaMaestro, the first feedforward model for metric house-scale panoramic reconstruction.
    \item We design a lightweight camera pose decoder for stronger pose estimation with large displacement.
    \item We propose a ERP panoramic data augmentation method for more pose-view pairs with existing data.
\end{itemize}

\section{Related Work}
\subsection{3D Foundation Models}
Feed-forward models have advanced significantly in point cloud reconstruction while predicting multiple 3D attributes, and become 3D Foundation Models for many downstream vision tasks.
The early DUSt3R \cite{wang2024dust3r} and MASt3R \cite{leroy2024grounding} predict a coupled scene representation but require further post-processing.
The following works expand them towards more pipelines \cite{duisterhof2025mast3r, murai2025mast3r, elflein2025light3r, pataki2025mp} and support extra input views \cite{cabon2025must3r, wang20243d, wang2025continuous}, but with limited quality compared to traditional optimization.

Recently, multi-view models such as VGGT \cite{wang2025vggt} have made excellent progress in 3D prediction, with many valuable efforts in faster reconstruction \cite{shen2025fastvggt}, more accurate localization \cite{tang2024mv, reloc3r} and longer sequences \cite{Yang_2025_Fast3R, deng2025vggtlongchunkitloop, yuan2026infinitevggt}.
Notably, $\pi^3$ \cite{wang2025pi3} employs a fully permutation-equivariant architecture and achieves higher robustness, while MapAnything \cite{keetha2025mapanything} further enables a broad range of 3D vision tasks in a single feed-forward pass, pushing such models to real-world applications.
Later, Depth Anything 3 \cite{depthanything3} shows that a single plain transformer is sufficient as a backbone, achieving minimal modeling.

Despite strong performance, the existing 3D foundation models are commonly based on pinhole cameras, leaving the challenge of panoramic 3D yet to be conquered.
In this paper, we aim to provide a panoramic 3D foundation model for future applications.
While many researches choose to implement special design for panoramas, we also hold the perspective \cite{depthanything3} that architectural specialization is not really necessary and uses a vanilla DINO as our backbone.

\subsection{Panoramic Depth Estimation}
With much larger view range than pinhole cameras, panoramic depth estimation serves as a promising technique for cost-efficient depth prediction method that may be used in autonomous driving and embodied AI systems.
Compared with standard perspective imagery, panoramic images, which are most commonly represented in the equirectangular projection (ERP), introduce strong non-uniform distortion and a periodic discontinuity at the left–right image boundary.
These properties encourages many unique design choices for panoramic depth estimation \cite{yan2022multi, shen2022panoformer, peng2022high, zioulis2018omnidepth, pintore2021slicenet, wang2020bifuse, cao2025panda, wang2024depth, bai2024glpanodepth, liu2024estimating, piccinelli2025unik3d}.

Specifically, Depth Any Camera (DAC) proposes a zero-shot metric depth estimation framework that extends a perspective-trained model to handle cameras with varying fields of view, such as fisheye and 360-degree cameras \cite{guo2025depth}, but the robustness is still limited by data.
To address the challenge of data scarcity, DA$^{2}$ \cite{li20252} uses a data curation engine to generate high-quality panoramic depth data from perspective images.
The recent DAP \cite{lin2025depth} further leverages a large-scale dataset created by combining public data, synthetic data, and real-world images to build a foundation model for panoramic metric depth estimation.

However, these models solely focus on monocular panoramic depth estimation that lacks known camera poses, making reconstruction using these depth priors impossible. This stresses the need for panoramic pose estimation.

\subsection{Multi-view Panoramas}
While multi-view 3D of pinhole cameras has been thoroughly discovered, researches into multi-view panoramas still have a long way to go.

Many recent methods focus directly on panoramic 3DGS reconstruction \cite{ren2025panosplatt3r, bai2025360, chen2025splatter}, achieving photorealistic reconstruction.
However, they mainly operate on 2-5 views and are no built to be geometry-aware.
In order to reconstruct longer sequences, others \cite{tu2024panopose, zheng2025scene, yun2022improving, wang2022bifuse++} uses dedicated pose estimation network for panoramas that either solve relative pose estimation between image pairs or directly process on scene-level.
Nevertheless, these methods still suffer from the requirements of large view overlap, leaving the problem of house-level reconstruction efficiency yet to be solved.

\section{Method}
\begin{figure}[!t]
  \centering
  \includegraphics[width=\textwidth]{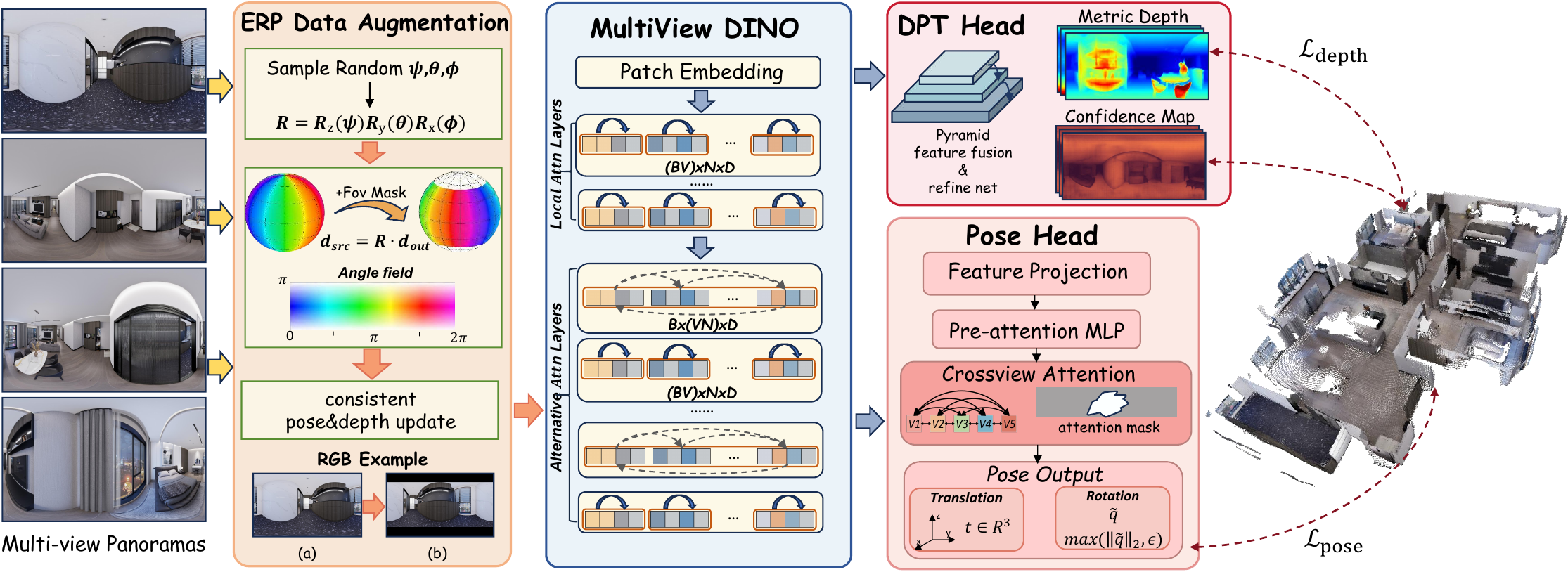}
  \caption{\textbf{The pipeline of CasaMaestro.} From left to right we illustrate ERP data augmentation, model backbone, functional heads and training objective.
  }
  \label{fig:pipe}
\end{figure}


\cref{fig:pipe} presents the overall architecture of the proposed CasaMaestro, which employs a multi-view DINO backbone to extract dense features from an arbitrary number of input panoramas. Leveraging these shared representations, the dedicated DPT and pose heads simultaneously estimate depth maps and camera extrinsics to facilitate scene point cloud back-projection. We elaborate on the specific design of these components below.

\subsection{Unified Backbone for Multi-view Representation Learning}
\label{sec:backbone_anyview}


To ensure an efficient model design, our backbone adopts the minimalist principle of utilizing a single plain ViT as in DepthAnything3 \cite{depthanything3}. We construct this foundation upon vanilla DINOv2 \cite{oquab2023dinov2} and facilitate multi-view reasoning via the input-adaptive cross-view self-attention mechanism, which is efficiently executed by reordering tokens during the forward pass.
The backbone is configured to process exclusively visual inputs as the goal is to solve indoor panoramic reconstruction without known camera extrinsics.


\paragraph{Tokenization.}
In our formulation, each individual panoramic image serves as a distinct view. Assuming an arbitrary number of such views $V$ for each sample, we represent the input images as
$\mathbf{I}\in\mathbb{R}^{B\times V\times 3\times H\times W}$.
We then apply a standard patch embedding to each view to obtain the patch tokens
$\mathbf{X}^{(0)} \in \mathbb{R}^{B\times V\times N\times D}$,
where $B$ stands for the batch size, $N$ represents the number of patches per view, and $D$ indicates the token dimension. We omit the class token for simplicity, although the formulation extends trivially if it is included. We use $\mathbf{X}^{(\ell)}$ to denote the token tensor produced after the $\ell$-th transformer block.


\paragraph{Intra-view self-attention.}
During the first $L_{\mathrm{local}}$ layers, we perform self-attention independently within each individual view. To implement this operation concretely, we reshape the tokens as $\mathcal{R}(\mathbf{X}^{(\ell)})\rightarrow \mathbb{R}^{(B V)\times N\times D}$ and subsequently apply a standard transformer block,

\begin{equation}
\mathbf{X}^{(\ell+1)} = \mathrm{TransBlock}_{\mathrm{local}}\!\left(\mathbf{X}^{(\ell)}\right),\qquad \ell < L_{\mathrm{local}}.
\label{eq:local_attn}
\end{equation}
By doing so, this initial stage effectively maintains the representation learning for each independent view and ensures that the early layers operate exactly like a standard transformer designed for single images.



\paragraph{Cross-view self-attention.}
Beginning at layer $L_{\mathrm{local}}$, the backbone alternates between local (intra-view) attention within individual views and global (cross-view) attention across multiple views. We achieve this alternation \textit{without} introducing any new attention layers. To implement this mechanism, we group tokens into distinct attention sequences. During the ``local step'', the model attends within each view as described in Eq.~\eqref{eq:local_attn}. Conversely, the ``global step'' attends across all views by flattening the view dimension directly into the token sequence. More specifically, for the global step, the tokens are reordered from $\mathbb{R}^{B\times V\times N\times D}$ to $\mathbb{R}^{B\times (V N)\times D}$. This resulting tensor is denoted as $\tilde{\mathbf{X}}^{(\ell)}$. Following this, we apply a standard transformer block over the sequence of length $VN$. We then reshape the resulting tensor back to its original dimensions,
\begin{align}
\tilde{\mathbf{X}}^{(\ell+1)} &= \mathrm{TransBlock}_{\mathrm{global}}\!\left(\tilde{\mathbf{X}}^{(\ell)}\right), \\
\mathbf{X}^{(\ell+1)} &= \mathrm{restore}\!\left(\tilde{\mathbf{X}}^{(\ell+1)}\right)\in\mathbb{R}^{B\times V\times N\times D}.
\label{eq:global_attn}
\end{align}

This backbone architecture effectively unifies local and global reasoning within a single transformer. The local attention mechanism maintains strong feature extraction for each individual view. Concurrently, the interleaved global steps allow tokens from distinct views to directly exchange information. This global interaction injects comprehensive context across the entire scene and significantly improves robustness even when viewpoints differ by large displacement.

We use standard DPT \cite{Ranftl2020, Ranftl2021} for per-view depth prediction, where multi-scale intermediate features are extracted for refinenet processing \cite{depthanything3}.
For camera pose head, we use the final output feature of backbone as input.
The backbone is camera-agnostic by default and supports an arbitrary number of input views with a single set of weights.

\subsection{Panoramic Camera Pose Decoder}

\paragraph{Problem setup.}
Given the features extracted for each individual view by the upstream backbone, we decode the camera pose parameters for panoramic / wide-FoV settings. We achieve this by employing a transformer decoder head that operates across multiple views. Let $\mathbf{F}\in\mathbb{R}^{B\times V\times C}$ denote the input features, where $C$ represents the feature dimension. The decoder output for each view $v$ is defined as follows.
\begin{equation}
\label{eq:poseheadout}
\hat{\mathbf{p}}_{b,v}=\big[\hat{\mathbf{t}}_{b,v},\ \hat{\mathbf{q}}_{b,v}\big]\in\mathbb{R}^{7},
\end{equation}
Here, $\hat{\mathbf{t}}_{b,v}\in\mathbb{R}^{3}$ represents the translation component, and $\hat{\mathbf{q}}_{b,v}\in\mathbb{R}^{4}$ denotes a unit quaternion rotation.




\paragraph{Motivation.}
A naive camera head can regress each view independently with an MLP, \textit{i.e.}, $\hat{\mathbf{p}}_{b,v}=g(\mathbf{f}_{b,v})$ \cite{depthanything3}. However, for panoramic  / wide-FoV inputs containing multiple images, this independent regression lacks explicit information exchange across views. While the upstream backbone already incorporates global attention, its primary role is to extract versatile features suitable for dense depth prediction. Camera pose estimation is an inherently global geometric task that requires aligning fully condensed representations from multiple perspectives. Forcing the backbone to resolve these specific viewpoint ambiguities could compromise its dense feature extraction capabilities. Therefore, to explicitly enforce global consistency and decouple this geometric alignment from the main backbone, we introduce a panoramic camera pose head. This head applies attention across views during the decoding stage, as illustrated in the bottom right of \cref{fig:pipe}. 
We provide validation of this design in \cref{sec:ablation}

\paragraph{Feature projection.}
We first project $\mathbf{F}$ to the model dimension $D$ with a linear layer followed by LayerNorm:
\begin{equation}
\mathbf{X}_0 = \mathrm{LN}\!\left(\mathbf{F}\mathbf{W}_p + \mathbf{b}_p\right)\in\mathbb{R}^{B\times V\times D}.
\label{eq:camdecpano_proj}
\end{equation}

\paragraph{Per-view refinement (pre-attention).}
Before cross-view attention, we apply a per-view MLP with residual connection and LayerNorm to improve token conditioning:
\begin{align}
\mathrm{MLP}(\mathbf{x}) &= \mathbf{W}_2\,\sigma(\mathbf{W}_1\mathbf{x}+\mathbf{b}_1)+\mathbf{b}_2,\ \sigma(\cdot)=\mathrm{GELU}(\cdot), \\
\mathbf{X}_1 &= \mathrm{LN}\!\left(\mathbf{X}_0 + \mathrm{MLP}(\mathbf{X}_0)\right).
\label{eq:camdecpano_premlp}
\end{align}
The MLP hidden dimension is set to $rD$ (with ratio $r$).

\paragraph{Cross-view self-attention.}
We then aggregate information across views through attention operated on the view dimension:
\begin{equation}
\mathbf{X}_2 = \mathrm{Attn}\!\left(\mathbf{X}_1;\ \mathbf{M}\right)\in\mathbb{R}^{B\times V\times D},
\label{eq:camdecpano_transenc}
\end{equation}
where $\mathbf{M}$ is an optional attention mask to handle invalid regions in real-world captures (\textit{e.g.}, camera poles or photographers needed to be masked out).
A final LayerNorm produces the fused tokens:
\begin{equation}
\mathbf{Z}=\mathrm{LN}(\mathbf{X}_2).
\label{eq:camdecpano_headln}
\end{equation}

\paragraph{Pose heads.}
Translation is regressed by a linear head:
\begin{equation}
\hat{\mathbf{t}}_{b,v} = \mathbf{Z}_{b,v}\mathbf{W}_t + \mathbf{b}_t \in \mathbb{R}^{3}.
\label{eq:camdecpano_t}
\end{equation}
Rotation is parameterized by a quaternion predicted by a linear head followed by explicit normalization:
\begin{align}
\tilde{\mathbf{q}}_{b,v} &= \mathbf{Z}_{b,v}\mathbf{W}_q + \mathbf{b}_q \in \mathbb{R}^{4}, \\
\hat{\mathbf{q}}_{b,v} &= \frac{\tilde{\mathbf{q}}_{b,v}}{\max\!\left(\lVert \tilde{\mathbf{q}}_{b,v}\rVert_2,\ \epsilon\right)}.
\label{eq:camdecpano_qnorm}
\end{align}
This enforces geometrically valid rotations and improves numerical stability during training.
For compatibility with the original DA3 behavior, if an external camera encoding $\mathbf{e}_{b,v}$ is provided, we optionally bypass the rotation regression and use its quaternion component $\hat{\mathbf{q}}_{b,v}=\mathrm{Norm}(\mathbf{e}_{b,v}[3{:}7])$.

While pinhole camera models may further estimate FOV values in $\mathbb{R}^{2}$, the standard panoramas come with fixed Field of View ($HFOV = 360^\circ,VFOV = 180^\circ$) and is hence free of prediction of such attributes, bringing a result of our pose decoder as in \cref{eq:poseheadout}.


\subsection{ERP Data Augmentation for Panoramas}
\label{sec:syn_rot_aug}


\paragraph{Motivation.}
In the training set from current available datasets, viewpoints within the same scene often exhibit little to no relative rotation (\textit{e.g.}, nearly identical yaw). This deficiency weakens the capacity of the model to develop reasoning capabilities regarding rotation, thereby hindering its generalization to diverse camera orientations. To address this limitation, we introduce an explicit rotation augmentation for equirectangular panoramic data equipped with paired RGB images, depth maps, and extrinsic parameters. This strategy produces additional relative rotations while strictly preserving the underlying 3D geometry.

\paragraph{Sampling random rotations.}
For each view in synthetic scenes, we sample a random rotation parameterized by yaw--pitch--roll:
\begin{align}
\psi &\sim \mathcal{U}(0,\ \psi_{\max}), \\
\theta &\sim \mathrm{clip}\big(\mathcal{N}(0,\sigma_{\theta}^{2}),\ -\theta_{\max},\ \theta_{\max}\big), \\
\phi &\sim \mathrm{clip}\big(\mathcal{N}(0,\sigma_{\phi}^{2}),\ -\phi_{\max},\ \phi_{\max}\big),
\end{align}
where $\psi$ is yaw, $\theta$ is pitch, and $\phi$ is roll.
We then construct a rotation matrix
\begin{equation}
\mathbf{R} = \mathbf{R}_{o_1}\mathbf{R}_{o_2}\mathbf{R}_{o_3},
\label{eq:aug_R_compose}
\end{equation}
where $(o_1,o_2,o_3)$ is a configurable multiplication order and $\mathbf{R}_x,\mathbf{R}_y,\mathbf{R}_z$ are standard axis-angle rotations.
In practice, $(o_1,o_2,o_3)$ is defined as $(z,y,x)$, and $\psi_{\max}, \theta_{\max}, \phi_{\max}$ are set to $\pi, \frac{\pi}{12}, \frac{\pi}{12}$ aligned with evaluation.

\paragraph{Equirectangular remapping.}
Given an equirectangular panorama of size $H\times W$, we associate each output pixel $(u,v)$ with spherical angles
\begin{equation}
\theta = 2\pi\left(\frac{u}{W}-\frac{1}{2}\right),\qquad
\varphi = \pi\left(\frac{1}{2}-\frac{v}{H}\right),
\label{eq:sph_angles}
\end{equation}
and convert to a 3D unit ray direction (using our panorama-ray convention):
\begin{equation}
\mathbf{d}_{\mathrm{out}}(\theta,\varphi)=
\begin{bmatrix}
\cos\varphi\ \sin\theta\\
-\sin\varphi\\
\cos\varphi\ \cos\theta
\end{bmatrix}.
\label{eq:ray_out}
\end{equation}
To synthesize a rotated panorama, we rotate rays and sample from the source panorama. With our implementation convention,
\begin{equation}
\mathbf{d}_{\mathrm{src}} = \mathbf{R}\ \mathbf{d}_{\mathrm{out}}.
\label{eq:ray_rotate}
\end{equation}
We then convert $\mathbf{d}_{\mathrm{src}}=[x_s,y_s,z_s]^\top$ back to spherical angles
\begin{equation}
\theta_s = \mathrm{atan2}(x_s, z_s),\qquad
\varphi_s = \arcsin\big(\mathrm{clip}(y_s,-1,1)\big),
\label{eq:sph_back}
\end{equation}
and finally obtain the sampling coordinates in the source image:
\begin{equation}
u_s = \left(\frac{\theta_s}{2\pi}+\frac{1}{2}\right)W,\qquad
v_s = \left(\frac{1}{2}+\frac{\varphi_s}{\pi}\right)H.
\label{eq:uv_back}
\end{equation}
We use bilinear interpolation and border wrap horizontally to respect the periodicity of panoramas.

\paragraph{Consistent pose update.}
Alongside warping the RGB and depth maps, we update the camera rotation in the extrinsic matrix to remain consistent with the augmented image.
Let $\mathbf{T}_{\mathrm{c2w}}\in\mathbb{R}^{4\times 4}$ be the camera-to-world transform. We update its rotation block by right-multiplication:
\begin{equation}
\mathbf{R}_{\mathrm{c2w}} \leftarrow \mathbf{R}_{\mathrm{c2w}}\ \mathbf{R}.
\label{eq:c2w_update}
\end{equation}

The same remapping is applied to both RGB and depth to preserve pixel-wise alignment after augmentation.
When a binary panorama mask is available, we apply it after the rotation warp (setting invalid pixels to zero), ensuring the masking operation remains consistent with the final augmented observations.

This augmentation injects controlled relative rotations into training scenes where native viewpoint rotations are limited, improving the model's robustness to orientation changes and encouraging cross-view reasoning under non-trivial camera rotations.

\section{Experiment}
\subsection{Implementation Details}
Our model is implemented using PyTorch framework and Adam optimizer.
CaseMaestro is trained for 20 epoches on $4\times$ NVIDIA A100 GPU with a per-GPU batchsize of 1 and a learning rate of 5e-5.
Input view number ranges from 20 to 50, and the DINO backbone is initialized from DA3~\cite{depthanything3}.
Evaluation is conducted on a single NVIDIA A100 GPU.
Training and evaluation resolution are set to 448 for single side with fixed aspect ratio as in DA3\cite{depthanything3}.
For pinhole methods, the data is split into pinhole images through ERP projection with $\frac{\pi}{2}$ yaw step (results are stable with different yaw steps as shown in \cref{fig:intro}).
All comparing methods are configured following their default settings.
CaseMaestro is trained on \textbf{Realsee3D} \cite{Li2025realsee3d_data} dataset, and we also evaluate the robustness of CaseMaestro by conducting zero-shot evaluation on other multi-view indoor panoramic datasets including \textbf{PanoSUNCG} \cite{wang2018self}, \textbf{The Habitat-Matterport 3D Research Dataset (HM3D)} \cite{ramakrishnan2021habitat} and \textbf{Replica} \cite{straub2019replica}.

\subsection{Quantitative Experiments}
\begin{table}[!t]
\centering
\caption{Quantitative pose metrics comparison on Realsee-Real. The $*$ denotes finetune or re-implementation on Realsee training set, and ${}^\circ$ indicates panoramic methods.}
\label{tab:pose_comparison_realsee_real}
\resizebox{\linewidth}{!}{
\begin{tabular}{lcccccc}
\toprule
Method & AUC@10 $\uparrow$ & AUC@20 $\uparrow$ & AUC@30 $\uparrow$ 
       & Rot. Mean $\downarrow$ & Trans. Mean $\downarrow$ & Pose Mean $\downarrow$ \\
\midrule
PanoPose*${}^\circ$\cite{tu2024panopose}  & 0.042 & 0.115 & 0.166 & 62.20 & 58.27 & 90.12 \\
SPR*${}^\circ$\cite{zheng2025scene}  & 0.086 & 0.145 & 0.191 & 59.43 & 56.12 & 86.23 \\
VGGT\cite{wang2025vggt} & 0.189 & 0.313 & 0.388 & 37.74 & 32.31 & 47.65 \\
VGGT-Finetune* & 0.276 & 0.379 & 0.411 & 30.12 & 10.23 & 38.42 \\
PI3\cite{wang2025pi3}  & 0.222 & 0.365 & 0.450 & 26.69 & 25.48 & 36.61 \\
PI3-Finetune* & 0.301 & 0.397 & 0.510 & 21.36 & 7.442 & 28.97 \\
DepthAnything3\cite{depthanything3}  & 0.083 & 0.173 & 0.235 & 54.57 & 49.79 & 69.51 \\
StreamVGGT\cite{streamVGGT} & 0.033 & 0.039 & 0.062 & 52.46 & 5.351 & 70.58 \\
InfiniteVGGT\cite{yuan2026infinitevggt} & 0.271 & 0.412 & 0.487 & 29.82 & 4.488 & 30.01 \\
CasaMaestro${}^\circ$(Ours)  & \textbf{0.727} & \textbf{0.859} & \textbf{0.903} & \textbf{2.608} & \textbf{2.132} & \textbf{3.225} \\
\bottomrule
\end{tabular}
}
\end{table}

\begin{table}[!t]
\centering
\caption{Quantitative pose metrics comparison on Realsee-Syn.}
\label{tab:pose_comparison_realsee_syn}
\resizebox{\linewidth}{!}{
\begin{tabular}{lcccccc}
\toprule
Method & AUC@10 $\uparrow$ & AUC@20 $\uparrow$ & AUC@30 $\uparrow$ 
       & Rot. Mean $\downarrow$ & Trans. Mean $\downarrow$ & Pose Mean $\downarrow$ \\
\midrule
PanoPose*${}^\circ$\cite{tu2024panopose}  & 0.082 & 0.135 & 0.179 & 57.44 & 57.39 & 88.04 \\
SPR*${}^\circ$\cite{zheng2025scene}  & 0.093 & 0.162 & 0.201 & 51.24 & 50.16 & 67.42 \\
VGGT\cite{wang2025vggt} & 0.118 & 0.226 & 0.292 & 48.93 & 42.99 & 61.49 \\
VGGT-Finetune* & 0.244 & 0.386 & 0.423 & 29.20 & 9.672 & 30.78 \\
PI3\cite{wang2025pi3}  & 0.237 & 0.347 & 0.410 & 38.82 & 21.77 & 28.65 \\
PI3-Finetune* & 0.310 & 0.412 & 0.507 & 23.43 & 5.320 & 25.31 \\
DepthAnything3\cite{depthanything3}  & 0.030 & 0.073 & 0.110 & 71.24 & 59.67 & 87.20 \\
StreamVGGT\cite{streamVGGT} & 0.007 & 0.007 & 0.010 & 57.84 & 11.56 & 102.0 \\
InfiniteVGGT\cite{yuan2026infinitevggt} & 0.068 & 0.196 & 0.288 & 61.37 & 6.991 & 61.83 \\
CasaMaestro${}^\circ$(Ours)  & \textbf{0.792} & \textbf{0.892} & \textbf{0.927} & \textbf{1.553} & \textbf{1.682} & \textbf{2.281} \\
\bottomrule
\end{tabular}
}
\end{table}

We conduct quantitative experiments on the sub-tasks of 3D point cloud reconstruction including camera pose estimation and depth estimation.

\paragraph{Camera Pose Estimation.}
We report the results of AUC and pose error of the Realsee dataset (test set) in \cref{tab:pose_comparison_realsee_real} and \cref{tab:pose_comparison_realsee_syn}, where pose error is defined as the maximum error of rotation error and translation error.
As shown, previous panoramic methods PanoPose and SPR rely highly on view overlap fail largely even when re-implemented on Realsee dataset, because these methods are designed for video streams with dense input frames.
Foundation models show relatively better quality due to large pre-training but are not stable as the metrics of both DepthAnything3 and InfiniteVGGT largely drop when it comes to synthetic scenes.
VGGT and Pi3 finetuned on Realsee training set is stable but fails to acquire significant improvements.
InfiniteVGGT and StreamVGGT both have low translation error, but still faces a large rotation error when applied to inputs with little overlap instead of video streams.
CasaMaestro stands the only one with the ability to correctly restore house structures, outperforming these methods with \textbf{84\%} and \textbf{119\%} improvements of the AUC30 metric than the best previous results in real-world and synthetic scenes with balanced rotation and translation error.

\paragraph{Depth Estimation.}
We first compare the depth estimation quality on Realsee dataset as shown in \cref{tab:depth_realsee}. Although PanoPose shows good quality, it is re-implemented on Realsee training set and common VFMs all provide zero-shot results with similar quality.
CasaMaestro significantly outperforms these models on Realsee dataset, 
yet while depth estimation could be data dependent and comparing with pinhole VFMs may not be fair, we further showcase the zero-shot depth estimation ability of CasaMaestro in three unseen datasets comparing only with panoramic depth estimation methods.
As shown in \cref{tab:depth_three}, even compared to dedicated single-view panoramic depth estimation methods or models that predict relative depth, the metric depth predicted by CasaMaestro is still superior, specifically an average \textbf{21.98\%} decrease in AbsRel compared to the best previous results, while no modules are intentionally designed for this task.

The above pose and depth estimation results demonstrate the robustness of CasaMaestro's panoramic 3D reconstruction ability across different scenes.

\begin{table}[!t]
\centering
\caption{Quantitative depth estimation comparison on Realsee dataset.}
\label{tab:depth_realsee}
\resizebox{\linewidth}{!}{
\begin{tabular}{l|ccc|ccc|ccc}
\toprule
& \multicolumn{9}{c}{Data Split} \\
Method 
& \multicolumn{3}{c}{Real-World}
& \multicolumn{3}{c}{Synthetic}
& \multicolumn{3}{c}{Overall} \\
\cmidrule(lr){2-4} \cmidrule(lr){5-7} \cmidrule(lr){8-10}
& AbsRel $\downarrow$ & RMSE $\downarrow$ & $\delta_1 \uparrow$
& AbsRel $\downarrow$ & RMSE $\downarrow$ & $\delta_1 \uparrow$
& AbsRel $\downarrow$ & RMSE $\downarrow$ & $\delta_1 \uparrow$ \\
\midrule
PanoPose*${}^\circ$\cite{tu2024panopose} & 0.143 & 0.504 & 0.801 & 0.156 & 0.319 & 0.773 & 0.155 & 0.336 & 0.775 \\
DAP${}^\circ$\cite{lin2025depth} & 0.144 & 0.530 & 0.809 & 0.188 & 0.406 & 0.723 & 0.184 & 0.417 & 0.731 \\
VGGT\cite{wang2025vggt} & 0.159 & 0.572 & 0.772 & 0.163 & 0.332 & 0.751 & 0.162 & 0.354 & 0.753 \\
InfiniteVGGT\cite{yuan2026infinitevggt} & 0.159 & 0.381 & 0.756 & 0.189 & 0.361 & 0.697 & 0.186 & 0.363 & 0.702 \\
Pi3\cite{wang2025pi3}     & 0.423 & 0.792 & 0.031 & 0.140 & 0.308 & 0.802 & 0.166 & 0.352 & 0.732 \\
DepthAnything3\cite{depthanything3}  & 0.203 & 0.617 & 0.705 & 0.153 & 0.325 & 0.745 & 0.157 & 0.351 & 0.741 \\
CasaMaestro${}^\circ$(Ours)  & \textbf{0.078} & \textbf{0.482} & \textbf{0.940}
             & \textbf{0.079} & \textbf{0.183} & \textbf{0.975}
             & \textbf{0.078} & \textbf{0.205} & \textbf{0.972} \\
\bottomrule
\end{tabular}
}
\end{table}

\begin{table}[!t]
\centering
\caption{Quantitative zero-shot depth estimation comparison.}
\label{tab:depth_three}
\resizebox{\linewidth}{!}{
\begin{tabular}{l|ccc|ccc|ccc}
\toprule
& \multicolumn{9}{c}{Dataset} \\
Method 
& \multicolumn{3}{c}{Replica}
& \multicolumn{3}{c}{PanoSUNCG}
& \multicolumn{3}{c}{HM3D} \\
\cmidrule(lr){2-4} \cmidrule(lr){5-7} \cmidrule(lr){8-10}
& AbsRel $\downarrow$ & RMSElog $\downarrow$ & $\delta_1 \uparrow$
& AbsRel $\downarrow$ & RMSElog $\downarrow$ & $\delta_1 \uparrow$
& AbsRel $\downarrow$ & RMSElog $\downarrow$ & $\delta_1 \uparrow$ \\
\midrule
\multicolumn{10}{l}{\colorbox{gray!20}{\textbf{Single-View Methods}}} \\
PanoFormer\cite{shen2022panoformer} & 0.076 & 0.129 & 0.945 & 0.054 & 0.122 & 0.978 & 0.151 & 0.184 & 0.742 \\
DA${}^2$\cite{li20252}     & 0.070 & 0.091 & 0.966 & 0.060 & 0.181 & 0.976 & 0.164 & 0.195 & 0.751 \\
DAP\cite{lin2025depth}          & 0.124 & 0.161 & 0.878 & 0.130 & 0.169 & 0.841 & 0.143 & 0.216 & 0.749 \\
\midrule
\multicolumn{10}{l}{\textbf{Multi-View Methods}} \\
PanoPose*\cite{tu2024panopose}     & 0.116 & 0.134 & 0.937 & 0.094 & 0.142 & 0.955 & 0.215 & 0.226 & 0.708 \\
CasaMaestro (Ours)  & \textbf{0.058} & \textbf{0.086} & \textbf{0.970}
             & \textbf{0.042} & \textbf{0.116} & \textbf{0.988}
             & \textbf{0.105} & \textbf{0.130} & \textbf{0.921} \\
\bottomrule
\end{tabular}
}
\end{table}

\subsection{Qualitative Comparison}
\begin{figure}[!t]
  \centering
  \includegraphics[width=\textwidth]{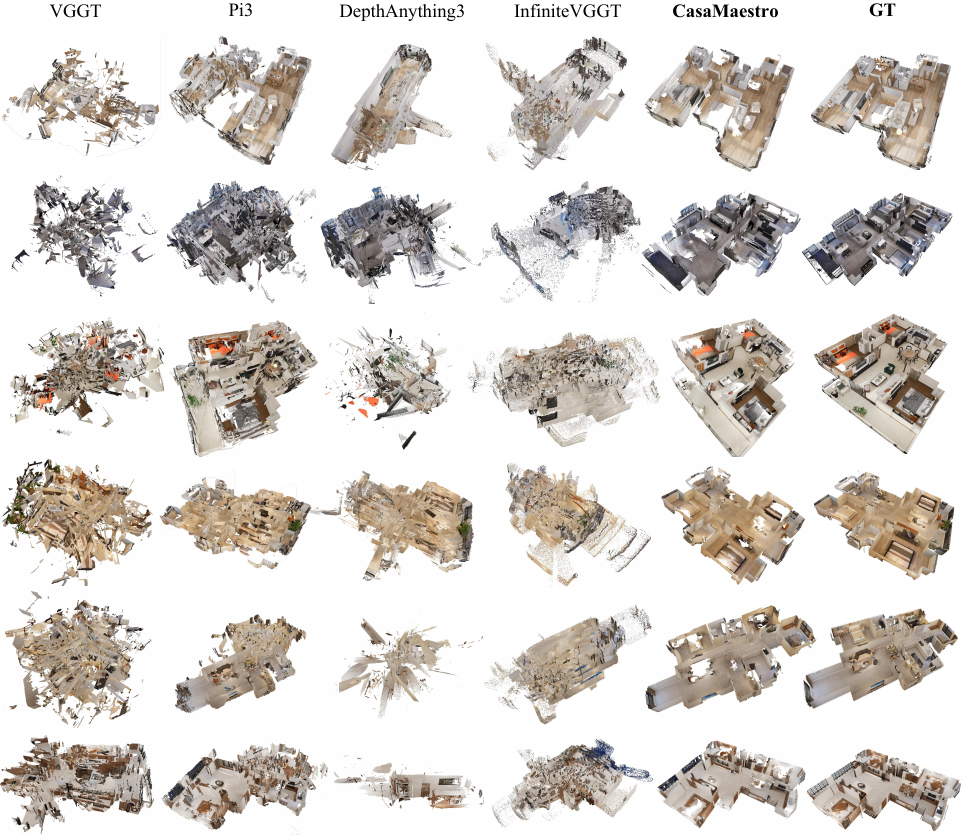}
  \caption{Quantitative comparison on Realsee-Syn. Upper 50\% points are filtered to remove the rooftop for visibility.}
  \label{fig:quan_syn}
\end{figure}

\begin{figure}[!t]
  \centering
  \includegraphics[width=\textwidth]{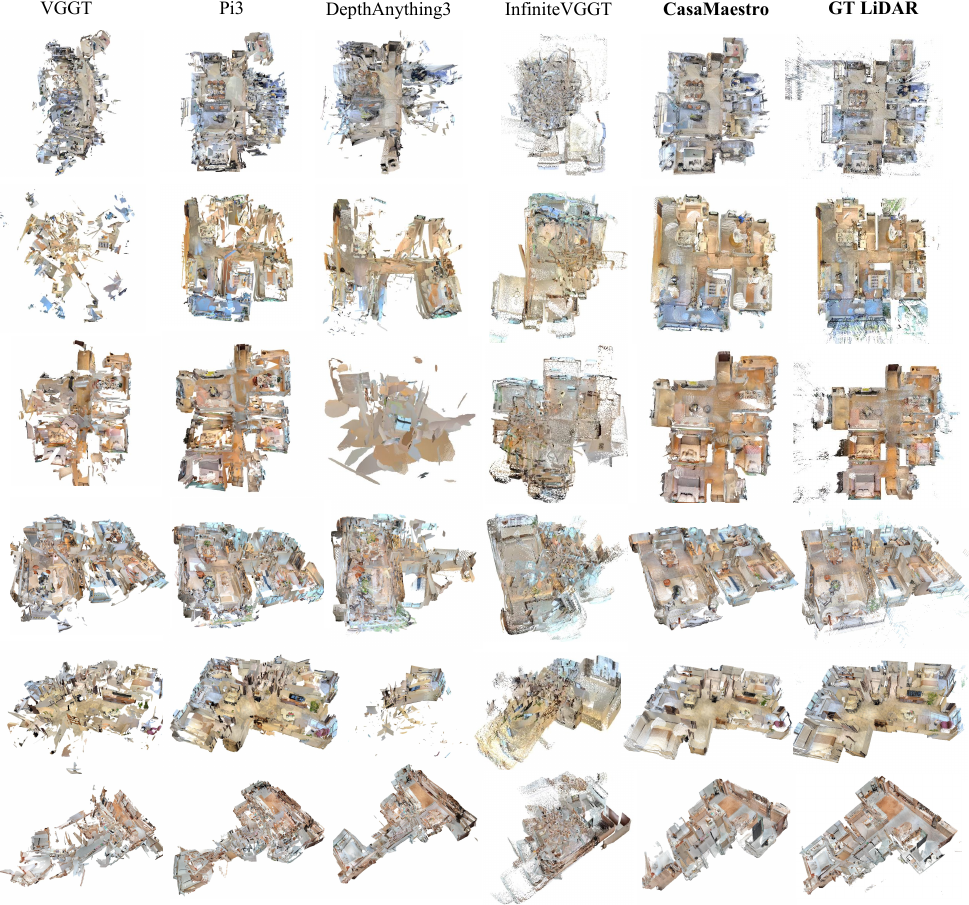}
  \caption{Quantitative comparison on Realsee-Real.}
  \label{fig:quan_real}
\end{figure}

We provide qualitative visual comparison in \cref{fig:quan_syn} and \cref{fig:quan_real}.
As shown, in circumstances of sparse viewpoints, all existing methods face severe errors and fail to generate reasonable reconstructions.
This result is in accord with metrics reported in \cref{tab:pose_comparison_realsee_real} and \cref{tab:pose_comparison_realsee_syn}.
Specifically, although InfiniteVGGT shows the least translation error among existing methods, the rotation error is high, which still leads to much pose error especially in synthetic scenes and the resulting reconstruction becomes messy.
Pi3 has more balanced rotation error and translation error, providing more reasonably reconstructed scenes, yet still creates much noise.
Meanwhile, VGGT and DepthAnything face more cases of complete corruption in certain scenes than others, standing for the high pose errors.
In the figures, CasaMaestro shows first-tier ability of accurately reconstructing houses with multiple rooms.

\subsection{Ablation Study}
\label{sec:ablation}

\begin{table}[t]
\centering
\caption{Ablation study with choices of key model components.}
\label{tab:ablation_modules}
\resizebox{\linewidth}{!}{
\begin{tabular}{c|ccccc|cccccc|ccc}
\toprule
\multicolumn{1}{c|}{+}
& \multicolumn{5}{c|}{Components} 
& \multicolumn{6}{c|}{Pose Metrics}
& \multicolumn{3}{c}{Depth Metrics} \\
\cmidrule(lr){1-1} \cmidrule(lr){2-6} \cmidrule(lr){7-12} \cmidrule(lr){13-15}
\# & DINO$_{\text{base}}$
& DINO$_{\text{large}}$
& PoseDec$_{\text{Linear}}$
& PoseDec$_{\text{Ours}}$
& ERP$_{\text{Aug.}}$
& AUC@10 $\uparrow$
& AUC@20 $\uparrow$
& AUC@30 $\uparrow$
& Rot. Mean $\downarrow$
& Trans. Mean $\downarrow$
& Pose Mean $\downarrow$
& AbsRel $\downarrow$
& RMSElog $\downarrow$
& $\delta_1 \uparrow$ \\
\midrule
\ding{172} & \checkmark &  & \checkmark &  &  
    & 0.302 & 0.454 & 0.568 & 24.39 & 17.85 & 28.11 & 0.092 & 0.110 & 0.920 \\
\ding{173} & \checkmark &  &  & \checkmark &  
    & 0.629 & 0.808 & 0.831 & 3.192 & 2.834 & 4.021 & 0.077 & 0.104 & 0.939 \\
\ding{174} & \checkmark &  & \checkmark &  & \checkmark
    & 0.512 & 0.651 & 0.772 & 2.175 & 2.375 & 2.989 & 0.111 & 0.125 & 0.970 \\
\ding{175} & \checkmark &  &  & \checkmark & \checkmark
    & 0.688 & 0.839 & 0.891
    & 2.547 & 2.404 & 3.380
    & 0.075 & 0.097 & 0.956 \\
\midrule
\ding{176} &  & \checkmark & \checkmark &  &  
    & 0.297 & 0.456 & 0.598 & 24.41 & 16.32 & 25.43 & 0.077 & 0.097 & 0.939 \\
\ding{177} &  & \checkmark &  & \checkmark &  
    & 0.707 & 0.829 & 0.871 & 2.102 & 2.175 & 2.837  & 0.076 & 0.105 & 0.973 \\
\ding{178} &  & \checkmark & \checkmark &  & \checkmark
    & 0.604 & 0.719 & 0.815 & 2.132 & 2.265 & 2.774 & 0.080 & 0.114 & 0.938 \\
\ding{179} &  & \checkmark &  & \checkmark & \checkmark
    & \textbf{0.792} & \textbf{0.892} & \textbf{0.927}
    & \textbf{1.553} & \textbf{1.682} & \textbf{2.281}
    & 0.078 & 0.103 & 0.975 \\
\bottomrule
\end{tabular}
}
\end{table}

\begin{figure}[!t] 
    \centering
    \begin{subfigure}[b]{0.48\textwidth}
        \centering
        \includegraphics[width=\textwidth]{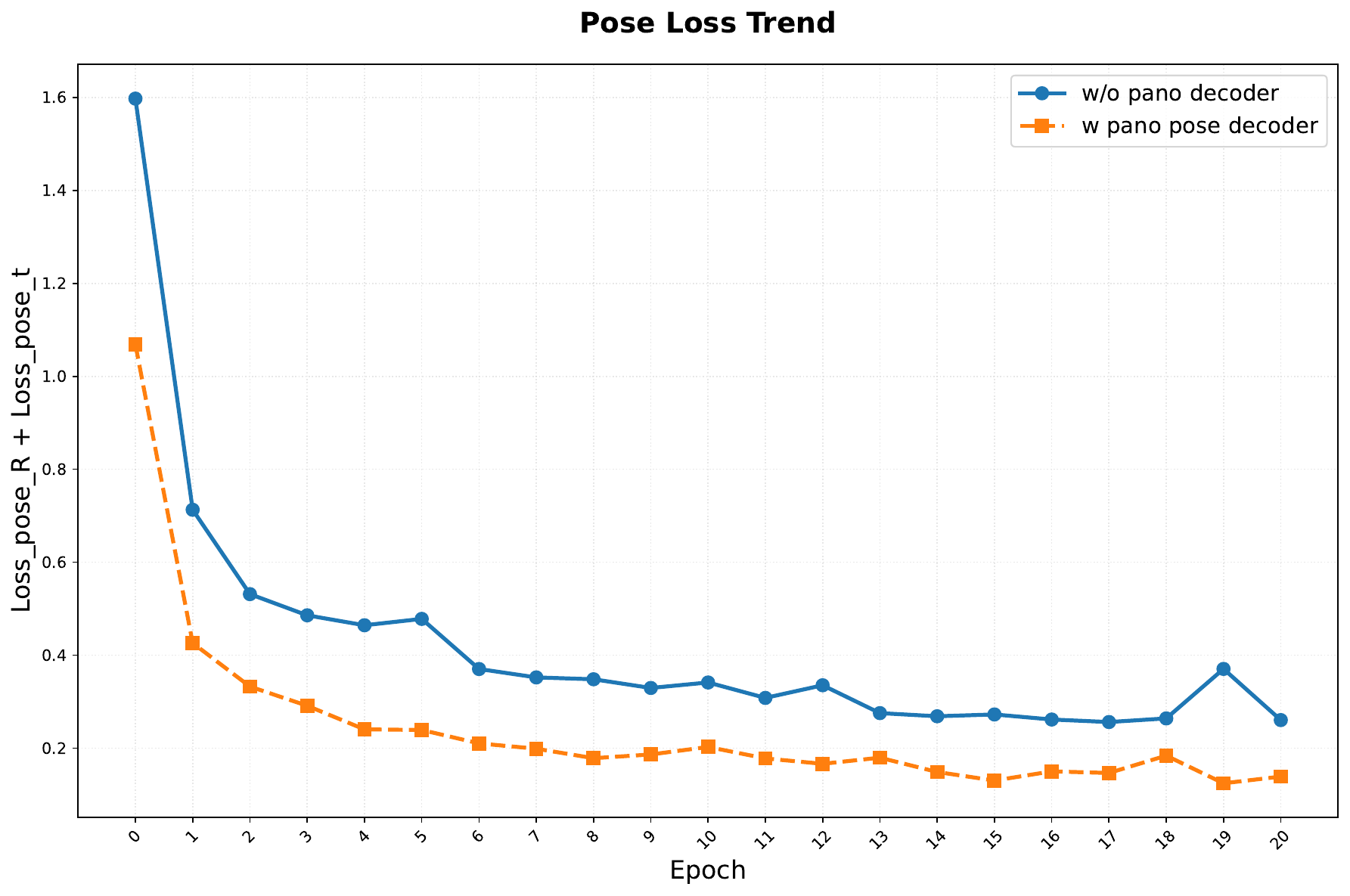}
        \caption{Ablation on pose decoder.}
        \label{fig:pose_loss}
    \end{subfigure}
    \hfill
    \begin{subfigure}[b]{0.48\textwidth}
        \centering
        \includegraphics[width=\textwidth]{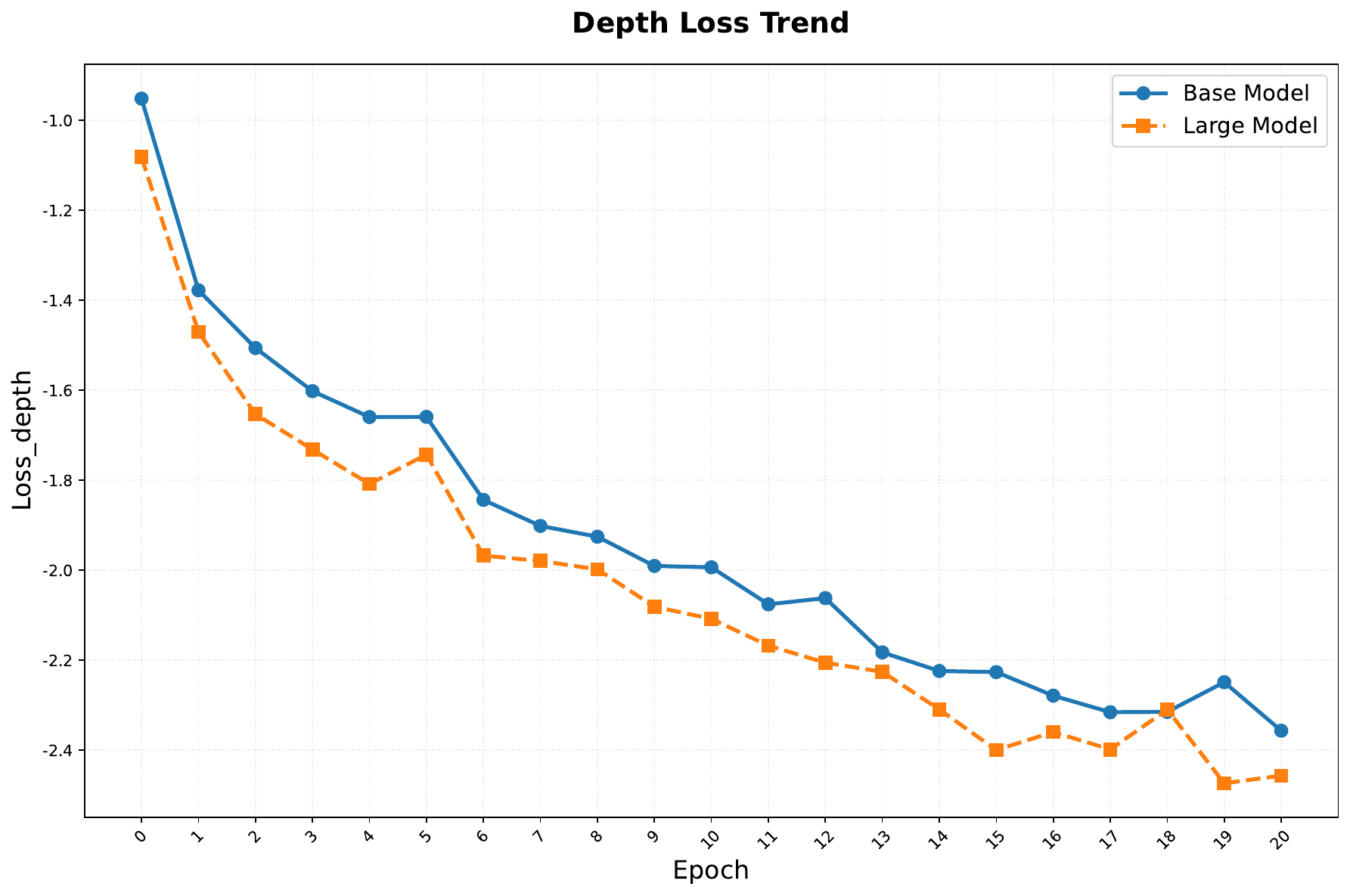}
        \caption{Ablation on model size.}
        \label{fig:depth_loss}
    \end{subfigure}
    \caption{Ablations with comparison of loss trends. Loss in epoch 0 means the average loss of the first epoch.}
    \label{fig:loss_trends}
\end{figure}

We conduct ablation to validate the rationality of our design.
The ablated components are model backbone, our panoramic camera pose decoder and ERP data augmentation strategy.

We provide comparison of final metrics in \cref{tab:ablation_modules}.
As reported, simply using a larger backbone does not show significant difference (\ding{172} \textit{v.s.} \ding{176}), which is in accordance with the trend of train-time depth loss shown in \cref{fig:depth_loss}, 
indicating that this problem can not be solved simply during feature extraction stage.
Using our panoramic camera pose decoder provides more improvements comparing with solely applying ERP data augmentation (\ding{173},\ding{177} \textit{v.s.} \ding{174},\ding{178}).
Meanwhile, as shown in \cref{fig:pose_loss}, during training, using the panoramic camera pose decoder will have much faster convergence than common linear decoder that takes only 3 epoches to reach the quality of 10+ epoches without it.
This shows that the extra attention applied to our proposed pose decoder indeed help with pose prediction instead of providing only metric improvements of the final results.

The above experiments demonstrate CasaMaestro's first-tier performance and the design rationality.
For more information including performance report and other experiments, please refer to the \textit{Supplementary Material}.

\section{Conclusion}
In this work, we present CasaMaestro, the first feedforward model that achieves extrinsic-free multi-view panoramic 3D reconstruction with sparse house-scale capture.
CasaMaestro significantly outperforms existing state-of-the-art models with an excellent ability to restore the house structure, while also showing a strong zero-shot ability of metric depth on unseen datasets.
The results prove that feedforward any-view 3D reconstruction needs not to be restricted to pinhole cameras, and the rise of more Vision Foundation Models built upon panoramas and fisheye cameras could provide a new revolution in various 3D tasks.

\section*{Acknowledgements} This work was supported in part by the Natural Science Foundation of China (Grant No.62503323).

%
%
\bibliographystyle{splncs04}
\bibliography{main}
\end{document}